\documentclass[conference]{IEEEtran}
\IEEEoverridecommandlockouts
\usepackage{cite}
\usepackage{amsmath,amssymb,amsfonts}
\usepackage{algorithmic}
\usepackage{graphicx}
\usepackage{textcomp}
\usepackage{booktabs} 
\usepackage{multirow} 
\usepackage{xcolor}
\usepackage{url}
\def\BibTeX{{\rm B\kern-.05em{\sc i\kern-.025em b}\kern-.08em
    T\kern-.1667em\lower.7ex\hbox{E}\kern-.125emX}}
\begin{document}

\title{Medical Imaging Fusing Vision Transformer: Laryngeal Cancer Screening with Explanation\\
}

\author{\IEEEauthorblockN{1\textsuperscript{st} Haiyang WANG}
\IEEEauthorblockA{\textit{Department of Electrical, Information and Bioengineering } \\
\textit{Polytechnic University of Milan}\\
Milan, Italy \\
haiyang.wang@polimi.it}
\and
\IEEEauthorblockN{2\textsuperscript{nd} Luca Mainardi}
\IEEEauthorblockA{\textit{Department of  of Electrical, Information and Bioengineering} \\
\textit{Polytechnic University of Milan}\\
Milan, Italy \\
luca.mainardi@polimi.it}
\and

}
\maketitle

\begin{abstract}
Early and timely screening of laryngeal cancer is crucial for improving clinical outcomes. In recent years, NBI endoscopy has become a standard diagnostic tool for the detection of laryngeal lesions. However, its effective use requires well-trained clinicians and the procedure is time-consuming and subject to interobserver variability. In this context, the application of artificial intelligence (AI) offers a promising solution to support clinical decision-making.

 In this work, we proposed applying transformer and attention mechanism for analyzing the narrow band imaging and distinguish benign and malignant lesions. Results show it has good classification performance with F1 (82.72\%), accuracy (82.33\%). In addition, the result of laryngeal cancer screening is explainable for clinicians. The explainability is utilizing the state of art segmentation method (MedSAM) to provide the useful pathological information area for clinicians. The proposed methodology fusing classification and segmentation  provides a translating on laryngeal cancer screening.

\end{abstract}


\section{Introduction}
Laryngeal cancer remains a significant global health concern, where early detection is paramount for effective treatment and improved survival rates\cite{hut2025laryngeal} \cite{han2025global}. In clinical practice, endoscopic examination, particularly with Narrow Band Imaging (NBI), serves as a primary diagnostic tool by enhancing the visualization of vascular patterns and mucosal structures \cite{yang2023narrative}\cite{jang2015past}. However, the interpretation of these images is inherently subjective, relying heavily on the clinician's expertise, and can be a labor-intensive process \cite{nocini2020updates} \cite{lam2025use}. This subjectivity and inefficiency can lead to diagnostic delays and inconsistencies.

Recent advances in Artificial Intelligence (AI), particularly in deep learning, present a compelling solution to these challenges by enabling automated, objective, and rapid medical image analysis \cite{rane2024artificial} \cite{bi2019artificial} \cite{tsuneki2022deep}. Machine learning applied to laryngeal cancer diagnosed could date back to  2002 by Ritchings et. al. \cite{ritchings2002pathological}, using artificial neural networks for pathological voice quality assessment for laryngeal cancer. Then, as NBI endoscopy \cite{de2017narrow} became a more standard diagnostic tool for laryngeal cancer diagnosed, reserchers started working more on NBI. In 2017, \cite{moccia2017confident} applied texture-based and SVM for laryngeal cancer diagnosis.
In 2019, Hao Xiong et. al.\cite{xiong2019computer}  applied convolutional neural networks (CNNs) 
 on laryngoscopic images for the first time.  Following this work, several studies have employed CNN-based approaches across various types of laryngeal imaging data, including hyperspectral imaging\cite{bengs2020spatio}. Although CNNs have been extensively applied in medical image analysis, the recently developed transformer architecture, renowned for its self-attention mechanism, may offers an alternative for capturing long-range dependencies in image data. 

In applying AI model in medical imaging, an understandable explanation is always significant in medical translating \cite{saw2025current}. In \cite{gupta2025interpretable} , Gupta et. al tried to applied LIME (Local Interpretable Model-Agnostic Explanations) as explanation for skin cancer. LIME explanations, though simplified and accessible, tend to be inconsistent unless refined through several rounds of feedback\cite{garreau2020explaining};
Grad-CAM (Gradient- weighted Class Activation Mapping) is another common method as CNN explanation. Such as in \cite{guluwadi2024enhancing} \cite{marmolejo2024numerical}\cite{seerala2020grad},  Grad-CAM provides heatmaps that localize important medical image areas. But, it requires access to gradients and designed for CNN; \cite{rahman2025enhanced} \cite{feng2025prediction} \cite{liu2025gadoxetic} applying SHAP (SHapley Additive 
exPlanations) as an explanation, it provides more stable and reliable estimates of feature contributions. But it sufferes complex permutations. In \cite{wang2024interpretable} radiomics is combined with Global Interpretation Index as an explaination from feature perspective. It bridges the gap between high-dimensional handcrafted descriptors and model-level explanation. Many existing explanation methods focus heavily on technical details and overlook the medical imaging context, producing explanations intended primarily for engineers rather than clinicians.

In our work, apart from applying AI to sort out laryngeal cancer patients, we provide an understandable explanation for clinicians.
We harness the potential of the transformer model and its inherent attention mechanism for the classification of laryngeal cancer from NBI endoscopic images. 
Our proposed framework is designed to effectively identify and focus on critical pathological features.  Meanwhile the state of art MedSAM \cite{ma2024segment} is applied to as the indicate potential suspicious pathological area as explanation. This could provide an understandable explanation for clinicians.

\section{Methodology}

\subsection{Dataset}\label{AA}
We used the publicly available CE-NBI dataset hosted on the Zenodo Repository (\url{https://zenodo.org/records/6674034}) \cite{zenodo}. This dataset comprises images extracted from contact endoscopy with narrow-band imaging (CE-NBI) video recordings of adult patients with suspected benign, premalignant, and malignant vocal fold lesions. All patients were examined at the Department of Otorhinolaryngology, Head and Neck Surgery, Magdeburg University Hospital, Germany, between January 1, 2015, and December 31, 2021.

\begin{figure}[ht]
\centering
\includegraphics[width=0.99\linewidth]{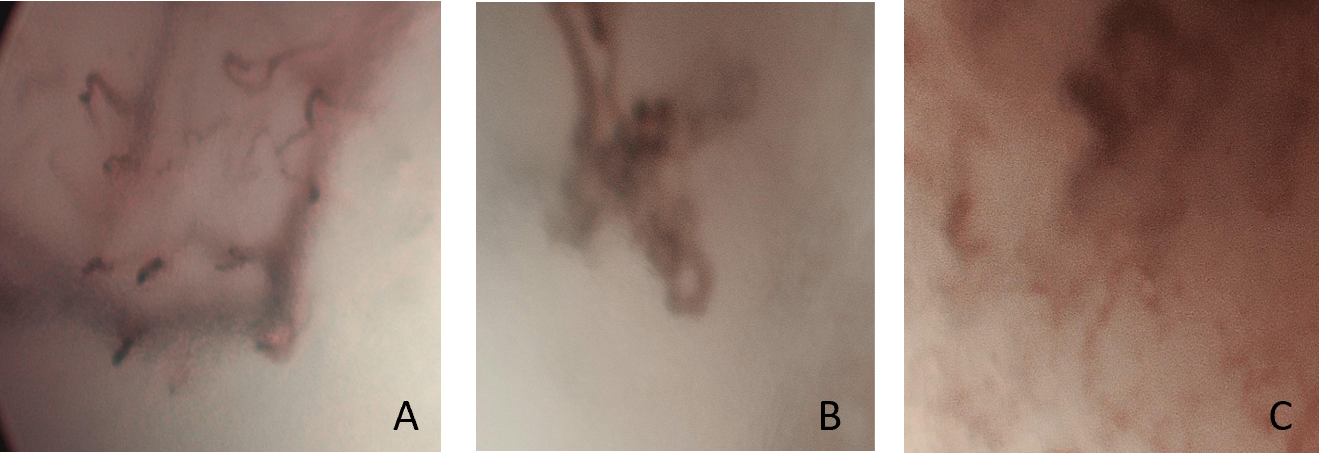}
\caption{\textbf{Samples of Malignant Laryngeal Cancer}. (A) Carcinoma in situ.(B) High grade dysplasia. (C) Squamous cell carcinoma.}
\label{Samples of Maligant}
\end{figure}

In total, the dataset includes 11,144 images from 210 adult patients, encompassing both benign and malignant cases. Each image is annotated with classification labels:  benign–malignant status \cite{zenodo}.
Samples of malignant laryngeal cancer are shown in Figure \ref{Samples of Maligant}.
Images are provided in JPG format with variable resolutions (e.g., 1280×1008, 1736×1080, 1842×1080, 720×544, and 868×540 pixels). Further details on dataset composition and structure are described  in \cite{esmaeili2023contact}.

\begin{figure*}[ht]
\centering
\includegraphics[width=0.8\linewidth]{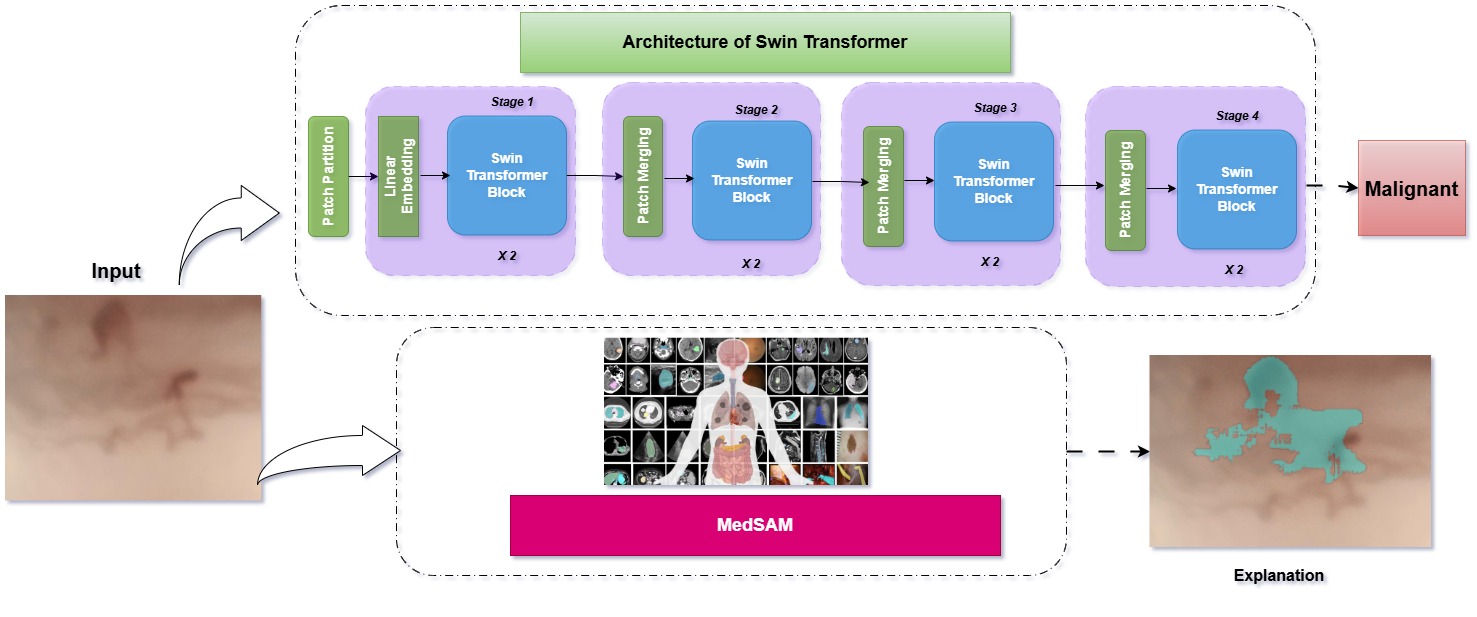}
\caption{ \textbf{Overview of the proposed framework combining Swin Transformer–based classification with MedSAM-based explanation.} The input NBI is first processed by a hierarchical Swin Transformer architecture consisting of patch partitioning, linear embedding, and four sequential stages with patch merging and repeated Swin Transformer blocks, producing a final classification output (e.g. malignant). In parallel, the same input image is fed into MedSAM to generate a visual explanation in the form of a segmentation or highlighted region, providing explanation for the model’s prediction.}
\label{scheme}
\end{figure*}

\subsection{Vision Transformer}
The Vision Transformer (ViT) is a neural architecture that transposes the standard Transformer, based on self-attention mechanisms and initially designed for sequential data in NLP, to process visual information by treating images as sequences of patches\cite{han2022survey}\cite{liu2021swin}.

ViT processes images by first partitioning the input into a sequence of non-overlapping patches. These patches are linearly projected into a lower-dimensional embedding space, forming the initial patch embeddings. To retain spatial information, learnable positional embeddings are added to these patch representations. This sequence, prepended with a special classification ([CLS]) token, is then fed into a standard Transformer encoder. The encoder comprises a stack of multi-headed self-attention layers, which enable global contextual integration by dynamically modeling dependencies between all patches, followed by position-wise feed-forward networks. For classification tasks, the final state of the [CLS] token serves as a comprehensive image representation, which is subsequently used for downstream tasks such as image classification.

\subsection{Data Preparation}

To develop our pipeline, the dataset is divided into training, validation and testing subsets, comprising $67.2\%$, $16.8\%$ and $16\%$ of the data, respectively. Splitting is performed at the patient level to ensure that images from the same patient were not shared between subsets, while maintaining a similar distribution of histopathology classes across both sets. The training set includes 7, 548 images, of which 5, 193 has benign lesions and 2, 355 has malignant lesions while the validaiton dataset is 1, 887 images with 1, 298 benign and 589 malignant. The testing set comprises 1, 709 images (Benign: 1, 166; Malignant: 543). 

 Each image is resized and normalized using the pretrained Swin processor corresponding to the model Swin Transformer \cite{swin_huggingface}. Here Swin-transformer is utizlied for its excellence in image classification than original Vision transformer. Figure \ref{scheme} describled the overview of the proposed framework.

\subsection{Patch Partition and Linear Embedding}

The input image $\mathbf{x} \in \mathbb{R}^{H \times W \times 3}$, where H is the height and W is the width, is divided into non-overlapping $4 \times 4$ patches. 
Each patch is flattened and linearly projected to a $C$-dimensional embedding using a learnable projection matrix:
\begin{equation}
    \mathbf{z}_0^i = \mathbf{x}_p^i \mathbf{E}, \quad \mathbf{E} \in \mathbb{R}^{(4^2 \cdot 3) \times C}.
\end{equation}
This patch embedding step is implemented internally in the Swin model as part of the patch embedding layer.

\subsection{Hierarchical Feature Representation}

Unlike the original Vision Transformer, which maintains a fixed spatial resolution throughout the network, the Swin Transformer \cite{liu2021swin} constructs a hierarchical feature representation by progressively merging patches across stages. At each stage, groups of $2 \times 2$ neighboring patches are concatenated and linearly projected, resulting in a doubling of the channel dimension:
\begin{equation}
\begin{aligned}
\hat{\mathbf{Z}} = \text{Linear}(&[\mathbf{z}(2i,2j), \mathbf{z}(2i+1,2j),  \\
&\mathbf{z}(2i,2j+1), \mathbf{z}(2i+1,2j+1)]).
\end{aligned}
\end{equation}
This operation reduces the spatial resolution by a factor of two while increasing the feature dimensionality, yielding a multi-scale representation analogous to that of convolutional neural networks.

\subsection{Window-Based Self-Attention and Shifted Windows}
Each Swin Transformer block applies self-attention within non-overlapping local windows of size $7 \times 7$. Given a feature map $\mathbf{X} \in \mathbb{R}^{h \times w \times C}$, the feature map is partitioned into $\frac{hw}{7^2}$ windows, and standard multi-head self-attention is performed independently within each window:
\begin{align}
\mathbf{Q}_h &= \mathbf{X}_w \mathbf{W}_h^Q, \quad
\mathbf{K}_h = \mathbf{X}_w \mathbf{W}_h^K, \quad
\mathbf{V}_h = \mathbf{X}_w \mathbf{W}_h^V, \\
\text{head}_h &= \operatorname{softmax}\!\left(
\frac{\mathbf{Q}_h \mathbf{K}_h^\top}{\sqrt{d_k}} + \mathbf{B}_w
\right)\mathbf{V}_h 
\end{align}

where $\mathbf{B}_w$ denotes a learnable relative position bias specific to each window.

To facilitate information exchange across windows, successive Swin Transformer blocks alternate between standard window partitioning and shifted window partitioning, in which windows are shifted by $(\frac{M}{2}, \frac{M}{2})$. This shifted window mechanism enables cross-window interactions while maintaining computational efficiency.

\subsection{Swin Transformer Block}

\begin{figure}[ht]
\centering
\includegraphics[width=0.8\linewidth]
{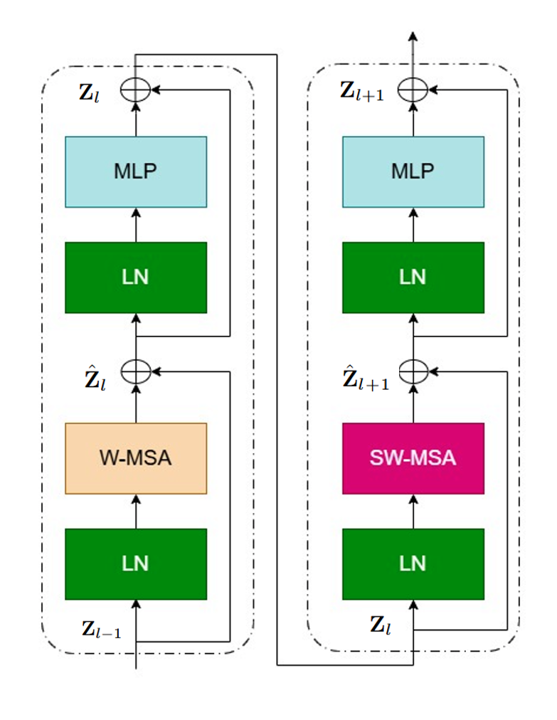}
\caption{\textbf{Architecture of a Swin Transformer block}. The first block applies Window-based Multi-Head Self-Attention (W-MSA), while the second employs Shifted Window Multi-Head Self-Attention (SW-MSA) to enable cross-window information exchange. In both blocks, Layer Normalization (LN) precedes the attention and MLP modules, and residual connections are applied after each sub-layer.}
\label{swin block}
\end{figure}

Each Swin Transformer block is illustrated in Figure \ref{swin block}, it is defined as:
\begin{align}
\hat{\mathbf{Z}}_l &= \text{W-MSA}(\text{LN}(\mathbf{Z}_{l-1})) + \mathbf{Z}_{l-1}, 
    \\
    \mathbf{Z}_l &= \text{MLP}(\text{LN}(\hat{\mathbf{Z}}_l)) + \hat{\mathbf{Z}}_l,
    \\
    \hat{\mathbf{Z}}_{l+1} &= \text{SW-MSA}(\text{LN}(\mathbf{Z}_{l})) + \mathbf{Z}_{l},
    \\
    \mathbf{Z}_{l+1} &= \text{MLP}(\text{LN}(\hat{\mathbf{Z}}_{l+1})) + \hat{\mathbf{Z}}_{l+1}
\end{align}

where W-MSA for window-based attention and SW-MSA for shifted-window attention. Layer Normalization (LN) is applied before each sub-layer (pre-norm structure), 
and MLP denotes a two-layer feed-forward network with GELU activation:
\begin{equation}
    \text{MLP}(\mathbf{X}) = \text{GELU}(\mathbf{X} \mathbf{W}_1 + \mathbf{b}_1)\mathbf{W}_2 + \mathbf{b}_2
\end{equation}

\subsection{Model Fine-Tuning and Optimization}
To accelerate training, the Swin backbone parameters $\mathbf{\theta}_{\text{swin}}$ were optionally frozen,
\[
\frac{\partial \mathcal{L}}{\partial \mathbf{\theta}_{\text{swin}}} = 0
\]
and only the classifier head parameters $\mathbf{\theta}_{\text{cls}}$ were optimized.

The classifier head is composed by a linear layer mapping the final pooled feature vector $\mathbf{z}_{\text{avg}} $ to two output classes:
\begin{equation}
    \mathbf{y} = \text{softmax}(\mathbf{z}_{\text{avg}} \mathbf{W}_{\text{cls}} + \mathbf{b}_{\text{cls}})
\end{equation}
where $\mathbf{W}_{\text{cls}} \in \mathbb{R}^{D \times 2}$ , $\mathbf{b}_{\text{cls}}$ is the bias term.

\subsection{Training and Evaluation}
During training, standard data augmentation techniques including random horizontal flipping, rotation, and color jittering are applied to improve generalization. Weighted random sampling is applied to address class imbalance. Training was performed for 20 epochs using the AdamW optimizer with a learning rate of $1\times10^{-4}$ and weight decay of $0.01$.
The model was evaluated using validation loss, with early stopping (patience = 3) and best-model checkpointing.

After training, the best-performing checkpoint was evaluated on the held-out test set (1,709 images).
Performance was reported using overall accuracy, precision, recall, F1-score, and the confusion matrix.

\subsection{Explanation}

MedSAM is a foundational model  designed to bridge the gap toward universal medical image segmentation \cite{ma2024segment}. The model is trained on a comprehensive dataset comprising 1.57 million paired medical images and corresponding segmentation masks, including endoscopy images. The dataset spans 10 distinct imaging modalities and covers more than 30 types of cancer. Extensive experiments across 146 tasks (86 internal and 60 external), encompassing diverse anatomies, pathologies, and imaging modalities, demonstrate that MedSAM consistently outperforms state-of-the-art segmentation foundation models \cite{nouman2024rethinking}.

In clinical applications of artificial intelligence, explainability is critical. However, providing meaningful explanations for AI predictions remains a significant challenge. To address this issue, we propose a novel approach that integrates segmentation as a form of explanation alongside classification tasks. Specifically, segmentation can visually highlight anatomical structures and pathological regions that are most relevant to the model’s decision, closely aligning with the regions clinicians routinely examine during diagnosis and treatment planning.

In this work, we employ MedSAM as the explanatory mechanism. MedSAM is based on a transformer backbone. Using segmentation outputs from the same transformer backbone provides a natural and consistent explanation for the classification decisions, ensuring architectural coherence between prediction and explanation.

\section{Results and discussion}

The performance of the proposed transformer-based model was quantitatively evaluated on a test set of 1,709 samples, with the detailed classification report presented in Table I. The results demonstrate a robust overall capability of the model in the defined classification task, while also revealing distinct inter-class performance variations that warrant further discussion.

\begin{figure}[ht]
\centering
\includegraphics[width=0.8\linewidth]{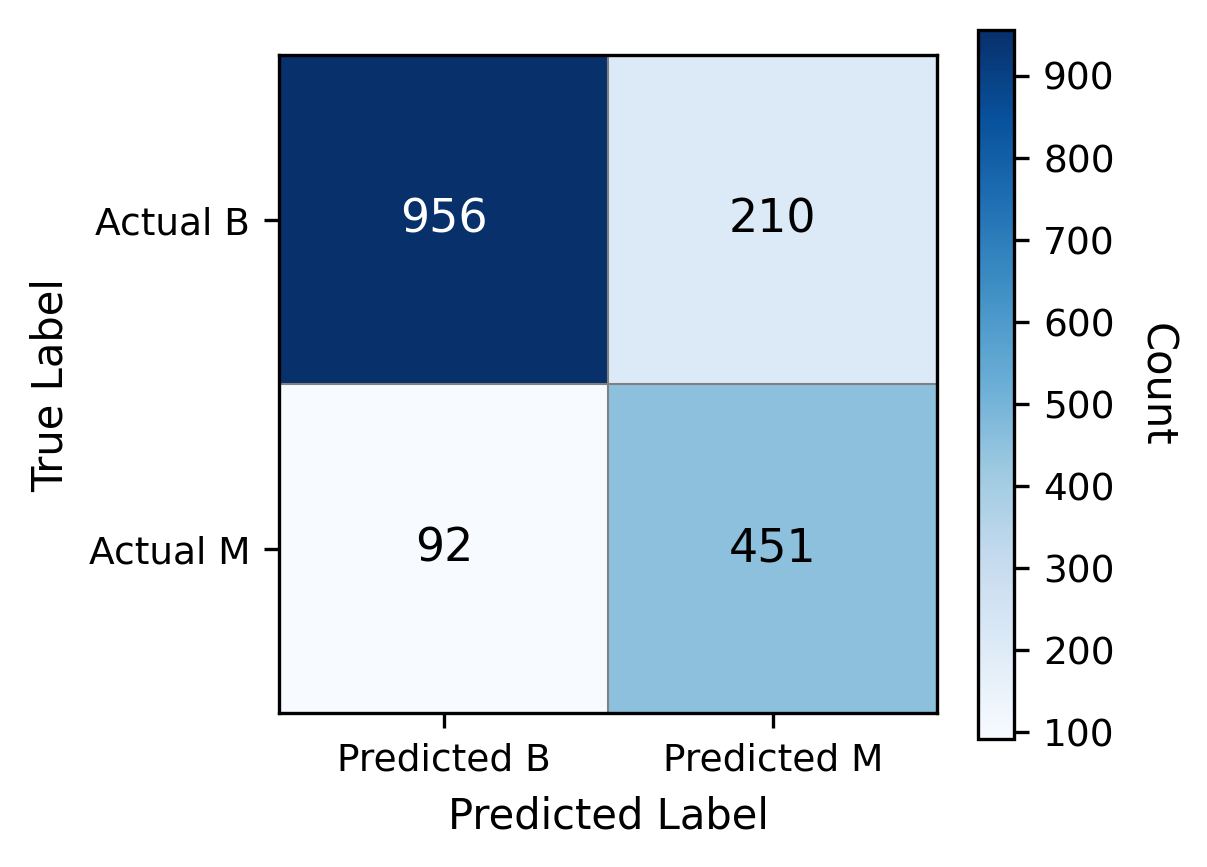}
\caption{Confusion Matrix of the Transformer Model}
\label{fig:confusion_matrix}
\end{figure}

Figure  \ref{fig:confusion_matrix} shows the confusion matrix of the proposed Transformer model on the test set. The model achieved 956 true negatives, 451 true positives, 92 false negatives, and 210 false positives, yielding an overall accuracy of 82.33\%. This is further corroborated by the weighted average F1-score of 82.72\% in Table. I, which provides a more reliable measure of overall performance given the class imbalance in the dataset (Support: Class B=1, 166, Class M=543). It demonstrated strong precision for malignant cases (91.22\%)—clinically valuable for reducing unnecessary biopsies. The close alignment between accuracy and the weighted average F1-score indicates that the model's performance is consistent across the distribution of the data.

\begin{table}[ht]
\caption{Transformer Classification Report}
\label{tab:classification_report}
\centering
\begin{tabular}{ccccc}
\bottomrule
\textbf{Class} & \textbf{Precision} & \textbf{Recall} & \textbf{F1-Score} & \textbf{Support} \\
\midrule
B (Benign) & 0.9122 & 0.8199 & 0.8636 & 1166  \\

M (Malignant) & 0.6823 & 0.8306 & 0.7492 & 543\\
\midrule
Overall Accuracy &  &  & 0.8233 & 1709 \\

Macro Avg & 0.7973 & 0.8252 & 0.8064 & 1709 \\

Weighted Avg & 0.8392 & 0.8233 & 0.8272 & 1709 \\
\bottomrule
\end{tabular}
\end{table}

\begin{figure}[ht]
\centering
\includegraphics[width=0.8\linewidth]{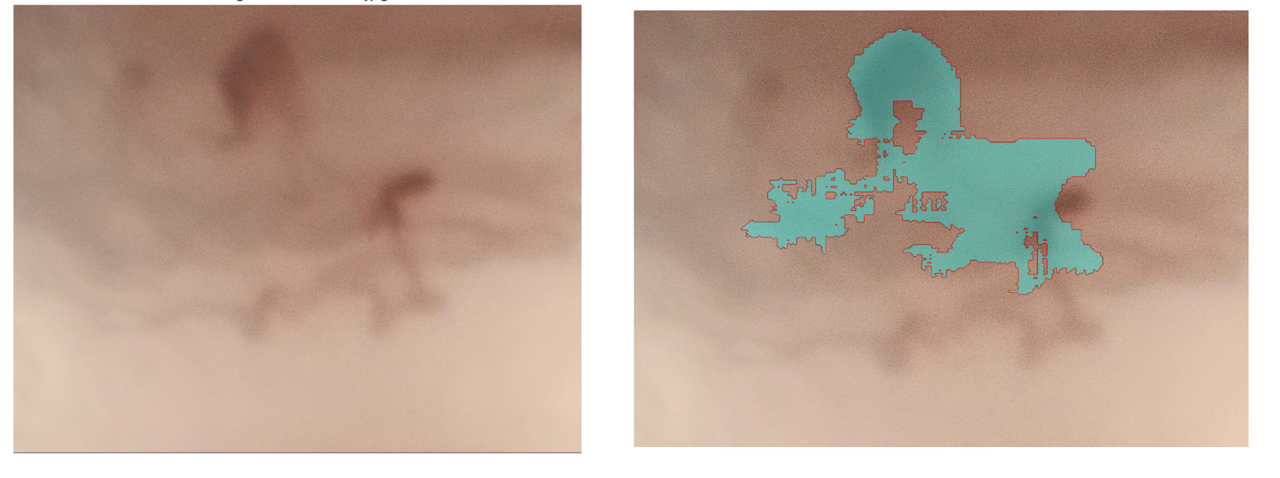}
\caption{Malignant Predication with Segmentation as Explanation}
\label{fig:Segmentation as Explanation}
\end{figure}

Fig. \ref{fig:Segmentation as Explanation}. illustrates a representative  image of a histopathologically confirmed malignant lesion alongside its corresponding segmentation area, which serves as a visual explanation of the model's decision-making process. The transparent green could cover most of the area of the pathological region. This proves MedSAM works with this laryngeal cancer dataset. There is a certain improvement space for the model. But it still indicates a relatively good explanation for AI model. Compared to conventional black-box classification approaches, the integration of segmentation as an intrinsic explanation mechanism enhances model transparency and trustworthiness — critical requirements for computer-aided diagnosis systems in laryngeal medical imaging diagnosis.

Experimental results confirm that our method achieves superior classification performance with understandable explanation. It provides a significant step towards translating AI-driven laryngeal cancer  diagnostic aids from research into clinical practice.


\section{Conclusion}

In our study, this paper has made a contribution in the field of
laryngeal cancer diagnosis with vision transformer. Transformer architecture, particularly with selective fine-tuning—markedly improves classification performance in laryngeal cancer detection on endoscopic images. (F1-score: 82.72\%; Accuracy: 82.33\%). Meanwhile, using MedSAM to achieve the area of the pathological region as the explantion to provide more instructive information. Here, we emphasized the importance of fusing classification and segmentation for AI in medical imaging. The findings underscore the transformative role of  transformer in medical image analysis and suggest promising directions for real-world deployment in laryngoscope. With continued refinement, such AI-assisted in laryngeal cancer could become invaluable.

\bibliographystyle{ieeetr}
\bibliography{references}



\end{document}